\pdfoutput=1

\documentclass[11pt]{article}

\usepackage[final]{acl}

\usepackage{times}
\usepackage{latexsym}

\usepackage[T1]{fontenc}

\usepackage[utf8]{inputenc}

\usepackage{microtype}

\usepackage{inconsolata}

\usepackage{graphicx}
\usepackage{amsmath, bm}
\usepackage{amsfonts}
\usepackage{url}
\usepackage{wrapfig,lipsum,booktabs} 
\usepackage{enumitem}
\usepackage{xcolor, colortbl}
\usepackage{pifont}
\usepackage{multirow,multicol}
\usepackage{mwe}
\usepackage{arydshln,booktabs, caption}
\usepackage{color,soul}
\usepackage{graphicx}

\DeclareMathOperator*{\argmax}{arg\,max}

\newcommand{\rX}[1]{\textcolor{red}{\ding{55}}{#1}}
\newcommand{\gC}[1]{\textcolor{green}{\ding{51}}{#1}}
\newcommand{\hlc}[2][yellow]{{%
    \colorlet{foo}{#1}%
    \sethlcolor{foo}\hl{#2}}%
}

\title{Light-weight Fine-tuning Method for Defending Adversarial Noise in Pre-trained Medical Vision-Language Models}

\newcommand{\affilsup}[1]{\rlap{\textsuperscript{\normalfont#1}}}

\author{Xu Han\affilsup{1} \quad Linghao Jin\affilsup{2} \quad Xuezhe Ma\affilsup{2} \quad Xiaofeng Liu\affilsup{1}\\
  $^1$\large{Yale University}\\
  $^2$\large{Information Sciences Institute, University of Southern California}\\
  \texttt{\{xu.han.xh365, xiaofeng.liu\}@yale.edu} \quad
  \texttt{\{linghaoj, xuezhema\}@isi.edu}
  }

\begin{document}
\maketitle
\begin{abstract}
Fine-tuning pre-trained Vision-Language Models (VLMs) has shown remarkable capabilities in medical image and textual depiction synergy. Nevertheless, many pre-training datasets are restricted by patient privacy concerns, potentially containing noise that can adversely affect downstream performance. Moreover, the growing reliance on multi-modal generation exacerbates this issue because of its susceptibility to adversarial attacks.
To investigate how VLMs trained on adversarial noisy data perform on downstream medical tasks, we first craft noisy upstream datasets using multi-modal adversarial attacks. Through our comprehensive analysis, we unveil that moderate noise enhances model robustness and transferability, but increasing noise levels negatively impact downstream task performance. To mitigate this issue, we propose rectify adversarial noise (RAN) framework, a recipe designed to effectively defend adversarial attacks and rectify the influence of upstream noise during fine-tuning.
\end{abstract}

\section{Introduction}

With the success of multi-modal learning~\citep{ngiam2011multimodal, tan2019lxmert, ramesh2021zeroshot, openai2024gpt4}, the availability of large medical Vision-Language Models (VLMs) has surged. Despite their potential, these models introduce considerable safety concerns. The pre-training datasets used on these VLMs are often inaccessible; maintaining data integrity becomes significantly challenging when scaling up. This issue is particularly pronounced in the healthcare domain, where data sensitivity and patient confidentiality limit its access. Consequently, they may contain imperceptible noise, which can adversely affect the model's generalization and transferability in downstream applications~\citep{havrilla2024understanding}, posing serious risks in medical contexts.

Additionally, as VLMs achieve remarkable success in generation tasks, there is a growing reliance on synthetic data~\citep{lu2024machine}. Examples include synthetic health reports~\citep{lee2023ehrsql}, medical instructions~\citep{belkadi2023generating}, and medical images~\citep{3ej9-e459-23}. This dependence on multi-modal generation exacerbates safety concerns, since models trained on such data are more susceptible to adversarial attacks~\citep{singh2024synthetic}. Adversaries can potentially compromise the entire system by subtly manipulating the most vulnerable modality.

\begin{figure}[!t]
\centering
\includegraphics[width=\linewidth]{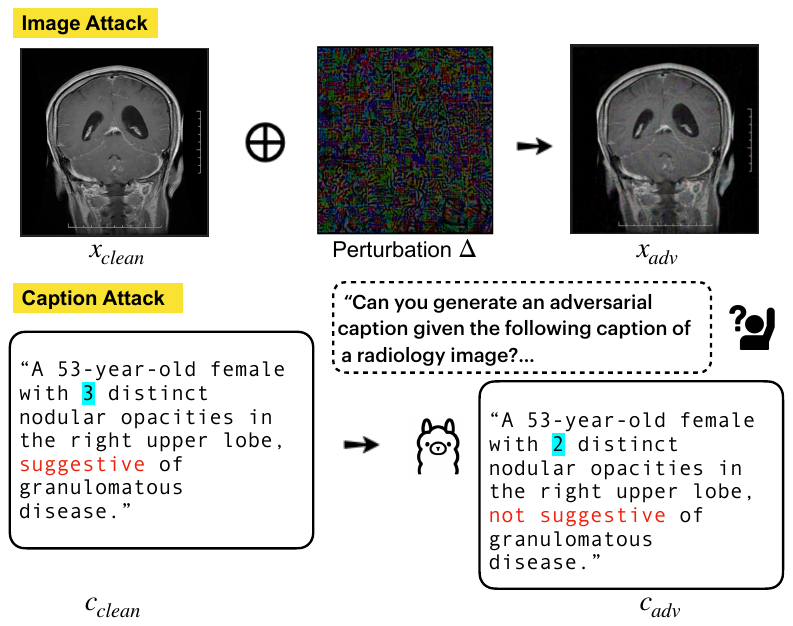}
\caption{The proposed multi-modal adversarial attack strategy.}
\label{fig:example_adv}
\vspace{-4mm}
\end{figure}

To remedy this issue, recent work has pivoted to understanding and mitigating noise introduced during pre-training~\citep{chen2024understanding}. Nonetheless, the robustness of VLMs pre-trained on perturbed adversarial samples remains unclear. This is particularly crucial in the healthcare context, where tasks like medical visual question answering (VQA) directly influence professionals' decisions regarding patient care.
In this paper, we focus on the following research:

\emph{Can we design a light-weight fine-tuning technique to alleviate the adverse effects introduced by adversarial noise in pre-trained Medical VLMs?}

Our main contributions can be summarized as follows:
\begin{enumerate}
    \item Towards modeling adversarial noise in pre-trained VLMs, we propose a novel multi-modal adversarial attacking strategy to perturb medical image-caption pairs (\autoref{fig:example_adv}), that effectively misleading victim VLMs (\S\ref{sec:noisy-data}).
    \vspace{-1mm}

    \item We introduce \textbf{R}ectify \textbf{A}dversarial \textbf{N}oise (RAN), a light-weight fine-tuning recipe to attenuate the effects of adversarial noise from pre-training (\S\ref{sec:mitigate-noise}).
    \vspace{-1mm}
    
    \item With empirical experiments, we pre-train \emph{noisy} medical VLMs using crafted adversarial data and evaluate the performance of such noisy models when fine-tuned on various downstream tasks including chest X-ray classification and medical VQA (\S\ref{sec:results}). 
    \vspace{-1mm}

\end{enumerate}

\section{Related Work}

\subsection{Medical Vision-Language Models}
Some recent efforts have broadened the scope of VLMs to the medical field for a variety of applications (i.e. medical report generation, medical VQA). For instance, MedViLL~\citep{Moon_2022} generates medical reports from images, aiding in clinical interpretation. PubMedCLIP~\citep{eslami-etal-2023-pubmedclip}, pre-trained on the ROCO dataset~\citep{10.1007/978-3-030-01364-6_20}, is fine-tuned for medical VQA to answer clinically relevant questions from visual inputs. BiomedCLIP~\citep{zhang2024biomedclip} adapts CLIP for biomedical applications, focusing on textual descriptions of medical images. LLaVa-Med~\citep{li2023llavamed} demonstrates advanced multi-modal conversational capabilities, making it highly popular in assisting with inquiries about biomedical images.
Such pre-trained models are typically trained on large-scale medical datasets, yet the quality of these datasets remains unexplored, and the robustness of the models has not been thoroughly evaluated.


\subsection{Adversarial Robustness}
Multi-modal VLMs are particularly vulnerable to adversarial attacks since perturbations can affect both visual and textual modalities. 
A number of general multi-modal adversarial attack strategies have been developed, targeting multiple tasks simultaneously~\citep{zhou2024revisiting, yin2024vlattack, zhao2023evaluating, cui2023robustness}. 
Recently, adversarial robustness has also garnered increasing attention in the medical sector. For example, \citet{thota2024demonstration} demonstrated how adversarial attacks on pathology images can mislead the Pathology Language-Image Pretraining (PLIP) model. Similar strategies have been employed in radiology to exploit the adversarial vulnerabilities of models used for medical imaging~\citep{Bortsova_2021, doi:10.1126/science.aaw4399}.

To improve robustness, \textit{adversarial training} has been shown to be one of the most effective approaches~\citep{shafahi2019adversarial, zhang2019theoretically, 9098740, xu2021robust, bai2021recent}. Training with adversarially perturbed samples enhances resistance to attacks but is time- and compute-intensive, especially for VLMs.
Recent studies have investigated parameter-efficient tuning techniques~\citep{mao2023understanding, ji2024advlora} 
to reduce the computational burden.
An alternative approach is \textit{adversarial purification}~\citep{nie2022diffusion, wang2022guided}, which uses diffusion models to transform adversarial examples back into clean representations.
These methods all attempt to enhance robustness during pre-training, we submerge the adversarial noise during fine-tuning in a privacy-protected~\citep{liu2022unsupervised} and light-weight black-box paradigm, assuming that pre-trained models are not always available. 


\subsection{Noise Learning and Robustness}
To address the challenges posed by noisy data during training, recent progress generally falls along two lines: robust model training and adapting clean pre-trained models on noisy (downstream) datasets.

Under the first taxonomy, techniques such as noise estimation~\citep{hendrycks2019using, Jiang_2019, xia2019t_revision, yao2021dual, goldberger2017training}, robust loss functions~\citep{ghosh2017robust, ma2020normalized} have been developed to mitigate the impact of noisy labels. Specifically, \citet{xue2022robust} proposes a robust co-training schema for medical image classification that iteratively filters out noisy samples. On the other hand, leveraging clean pre-trained models to adapt to noisy downstream datasets has proven to be both practical and efficient. Under this paradigm, effective fine-tuning strategies have been explored to enhance model robustness against noisy data~\citep{ wu2022noisytune, zhang2022pats}.

\begin{figure*}[t]
\centering
\includegraphics[width=\textwidth]{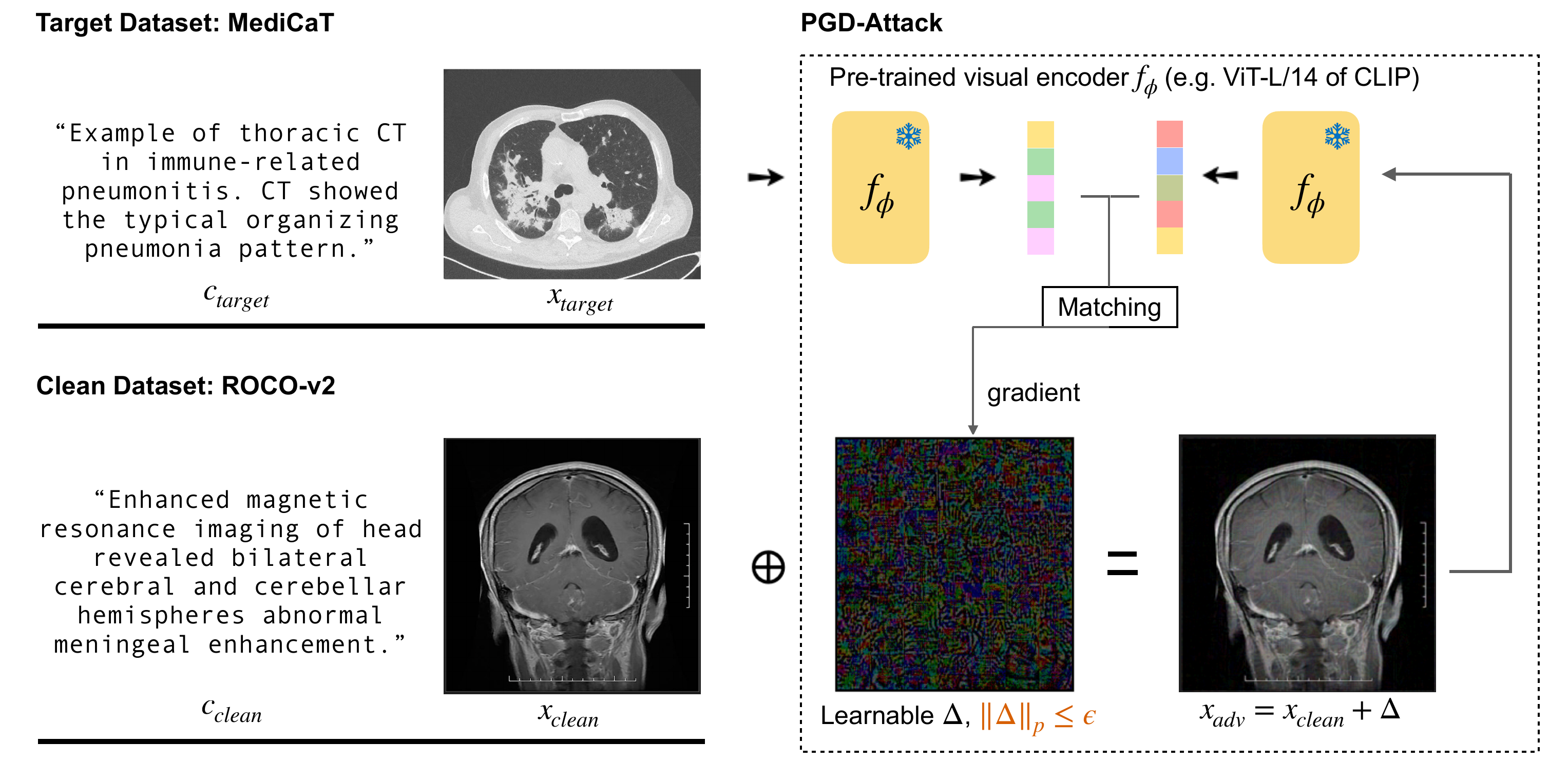}
\caption{\textbf{Pipeline of our radiology image attacking strategy}. We select a target image-caption pair \((x_\text{target}, c_\text{target})\) from \textsc{MediCaT} and a clean pair \((x_\text{clean}, c_\text{clean})\) from \textsc{ROCOv2}. $x_\text{target}$. Images are transformed to embeddings by pre-trained visual encoder $f_\phi$. Adversarial example $x_\text{adv}$ is generated by PGD (Eq \ref{eq:2}) iteratively and we denote the adversarial noise as $\Delta$. As formulated in Eq \ref{eq:1}, we optimize the $\Delta$ to maximize the similarity between $x_\text{target}$ and $x_\text{adv}$; the perturbation $\Delta$ is also limited by $\|\Delta\|\leq \epsilon$.}
\label{fig:adv_img}
\end{figure*}

Until recently, \citet{chen2024understanding} introduces \textit{noisy model learning}, which focuses on the effect of pre-training label noise on downstream. To our knowledge, no previous research has qualitatively assessed the effects of adversarial noise during pre-training on downstream tasks, particularly focusing on multi-modal noise in the medical domain.


\section{Methods}

In this section, we first introduce our novel multi-modal adversarial attack strategies designed to create noisy medical image-caption datasets, which will be used to train CLIP models. 
Then, we introduce RAN fine-tuning to alleviate the effect of such pre-trained noisy medical models on downstream classification tasks.


\subsection{Adversarial Noisy Dataset Generation}
\label{sec:noisy-data}
\paragraph{Notation} Consider a clean dataset \( D_\text{clean} = \{(x^i, c^i)\}_{i=1}^N \) for upstream training, where \( N \) is the total number of samples in the dataset. Each tuple includes a training image \(x^i\), its corresponding textual caption \( c^i \). 
Our goal is to generate a noisy dataset \(D_\text{noisy}\) to train a noisy CLIP model \(M_\text{noisy}\). 
We use noise ratio \(\gamma\) to denote the percentage of noisy samples in \(\hat{D}_\text{noisy} \). Each noisy sample \((x^i_\text{adv}, c^i)\) or \((x^i, c^i_\text{adv})\) is created through one of the following adversarial methods: \textit{image attack} or \textit{caption attack}.

\paragraph{Adversarial Image Attack.}
To craft adversarial images $x_{\text{adv}}$ from clean images $x_{\text{clean}}$ that can deceive \textit{victim} models, we use an image encoder $f_\phi$ from a publicly accessible model, i.e., ViT-L/14 of pre-trained CLIP models, as the surrogate model. Considering VLMs may be unreliable for optimizing cross-modality similarity~\citep{zhao2023evaluating}, we select a target image $x_{\text{target}}$ instead of a target caption $c_{\text{target}}$ to guide the generation of $x_{\text{adv}}$, ensuring $x_{\text{target}}$ and $x_{\text{clean}}$ comes from different data distribution. The adversary aims $x_{\text{adv}}$ to resemble $x_{\text{target}}$ through human imperceptible perturbations:


\begin{equation*}\label{eq:1}
\argmax_{\|x_{\text{clean}} - x_{\text{adv}}\|_p \leq \epsilon} f_{\phi}(x_{\text{adv}})^\top f_{\phi}(x_{\text{target}})
\end{equation*}

\begin{figure*}[t]
\centering
\includegraphics[width=0.8\textwidth]{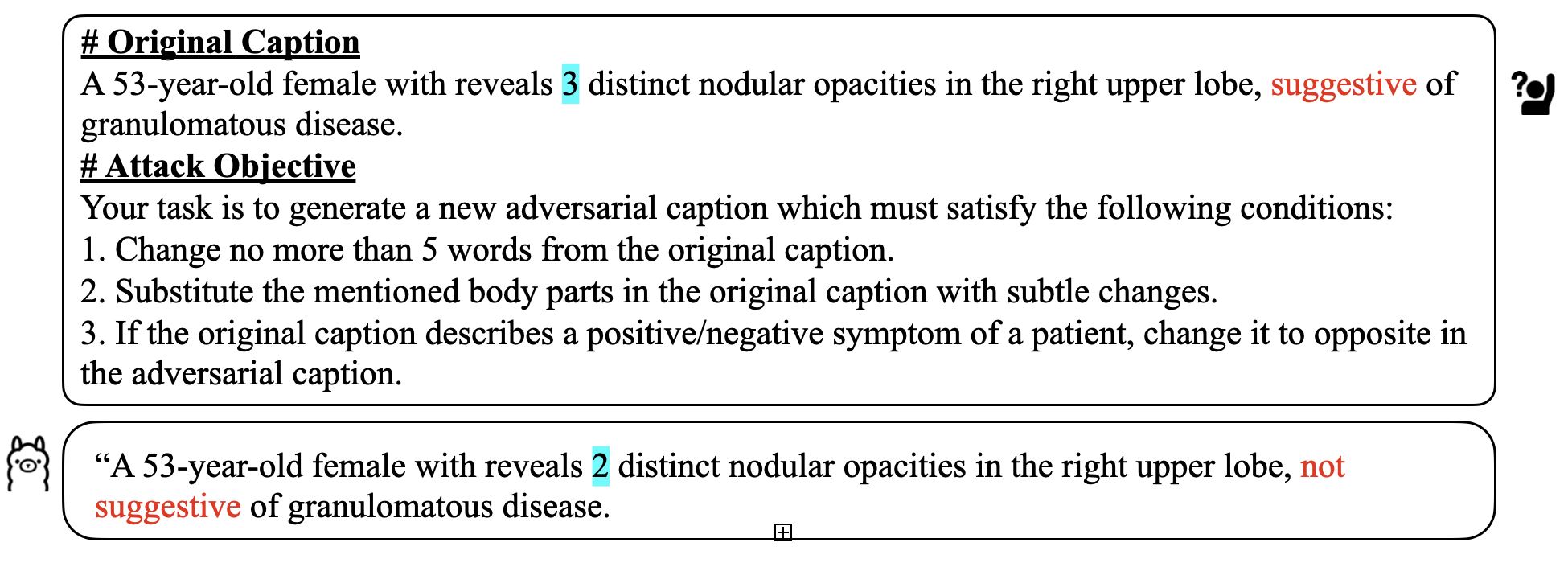}
\caption{Our proposed \textbf{prompt to generate an adversarial caption} of a corresponding radiology image. \\ \hlc[cyan!20]{Highlighted} are "body part" designed to change by prompt; \textcolor{red}{Words} are changed to opposite by prompt.}
\label{fig:prompt}
\end{figure*}

We utilize projected gradient descent (PGD)~\citep{madry2019deep} to address the constrained optimization problem presented. 
PGD iteratively applies gradient ascent on $x_{\text{adv}}$ to maximize the cross-entropy loss $\mathcal{L}$. Each iteration is characterized as
\begin{equation*}\label{eq:2}
x^{(t+1)} = \Pi \left( x^{(t)} + \alpha \cdot \text{sign} \left( \nabla_x \mathcal{L} (\theta, x^{(t)}, c) \right) \right)
\end{equation*}
Here, \( x^0=x_\text{clean}, \Pi \) is a projection to guarantee adversarial perturbation remains within the acceptable limits. The process to generate adversarial image samples is illustrated in \autoref{fig:adv_img}.

\paragraph{Adversarial Caption Attack.}
Prompt-based adversarial attacks are capable of independently and effectively discovering the weaknesses of a victim LLM~\citep{xu2023llm}. Given \( (x^i, c^i) \) in the pre-train dataset \(D_\text{clean}\), $c^i$ represents a caption describing a radiology image $x_i$ in our case. We alter the captions $c^i$ with \textsc{Llama3-8b}\footnote{\url{https://ai.meta.com/blog/meta-llama-3/}} using a prompt 
that incorporates three attack objectives, as depicted in \autoref{fig:prompt}. These objectives generate adversarial captions $c_\text{adv}$ that fail to accurately describe the corresponding radiology image, with few key words (i.e. body parts, symptom) adjustment. 




\begin{figure*}[t]
\centering
\includegraphics[width=\textwidth]{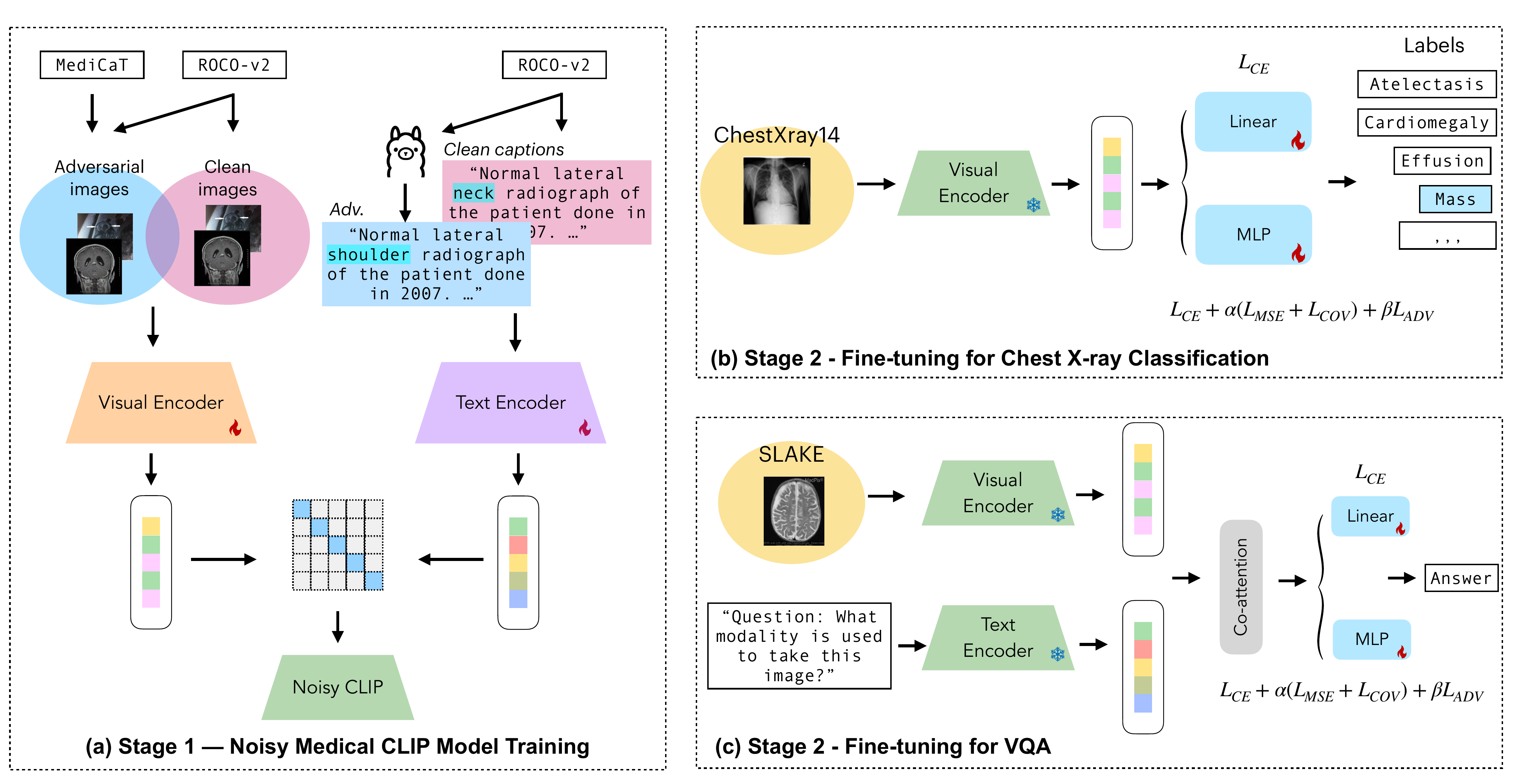}
\caption{Illustration of \textbf{(a)} training a noisy model with a combination of adversarial and clean data. The trained noisy model is then fine-tuned on \textbf{(b)} chest x-ray classification task and \textbf{(c)} medical VQA task. In \textbf{(c)}, we employ a co-attention module to fuse textual and visual features before feeding into a classifier. The classifier can be either a linear classification head or an MLP. }
\vspace{-2mm}
\label{fig:architecture}
\end{figure*}

\subsection{RAN: Rectify Adversarial Noise}
\label{sec:mitigate-noise}
Our fine-tuning objective consists of three components: a covariance loss to attenuate noise impact, a consistency loss, and an adversarial loss to defend adversarial attack in classification tasks.



\paragraph{Covariance Loss.} 
\citet{chen2024understanding} observed that introducing noise diminishes the top dominant singular values of the pre-trained features, leading to reduced transferability. Building on this insight, we transform pre-trained features \(\mathcal{F}\) into a new feature space \( \mathcal{Z} \) using multi-layer perceptron (MLP) with covariance regularization term \citep{bardes2022vicreg} to rectify effects of the introduced noise :

\begin{equation*}\label{eq:4}
\mathcal{L}_{\text{COV}} = \frac{1}{D} \sum_{i \neq j} [C(\mathcal{Z})]_{i,j}^2, \quad 
\end{equation*}
where \(C(\mathcal{Z})\) is defined as the covariance matrix of transformed features \( \mathcal{Z}\):
\begin{equation*}
    C(\mathcal{Z}) = \frac{1}{n - 1} \sum_{i=1}^n (z_i - \bar{z})(z_i - \bar{z})^T, \bar{z} = \frac{1}{n} \sum_{i=1}^n z_i
\end{equation*}

By minimizing the off-diagonal coefficients of \( C(\mathcal{Z}) \) to approach zero, we encourage the features to encode discriminative information.

\paragraph{Consistency Loss.} To maintain the pre-trained knowledge unchanged, we use a mean-square-error (MSE) loss between the normalized features \(\mathcal{F}\)  and \( \mathcal{Z} \):
\begin{equation*}
\mathcal{L}_{\text{MSE}} = \left\| \hat{\mathcal{F}} - \hat{\mathcal{Z}} \right\|_2^2. \label{eq:3}
\end{equation*}
Here, \( \hat{\mathcal{F}}= \frac{\mathcal{F}}{\|\mathcal{F}\|_2} \) and \( \hat{\mathcal{Z}}= \frac{\mathcal{Z}}{\|\mathcal{Z}\|_2} \). This objective aids in transferring the pre-trained knowledge to the transformed features \( Z \).



\paragraph{Adversarial Loss.}
Cross-entropy loss often struggles to distinguish adversarial samples in the feature space because it does not explicitly enforce a robust margin between learned classes~\citep{xia2022tightening}. 

Given \((x_i, y_i)\) in a classification task, $f_i$ denotes the features of $x_i$ from pre-trained noisy model $M_\text{noisy}$.
To address this issue and enhance the robustness of the trained classifier against adversarial attacks, we introduce a constraint to maximize: 1) the distance between the features $f_i$ of a given class $y_i$ and the learned centroids of other classes, and 2) the separation between the learned centroids of different classes:
\begin{equation*}
\mathcal{L}_{\text{ADV}} =-\frac{1}{D}\sum_{i=1}^D \text{dist}(f_i) + \arccos(c_{y_i} \cdot c_j), \quad 
\end{equation*}
\vspace{-3mm}
\begin{equation*}
    \text{dist}(f_i) = \frac{1}{k-1} \sum_{j \neq y_i} ^{k-1}\left\| f_i - c_{y_i} \right\|
\end{equation*}


The $c_{y_i}$ denotes the $y_i$th class center of features. $\text{dist}(f_i)$ encourages the $f_i$ to be away from wrong classes’ centroids. 
$\mathcal{L}_{\text{ADV}}$ enables the decision margins between the centroids of the classes to be separated sufficiently to prevent the overlapping of features from different classes.

The overall loss function for downstream classification tasks becomes :
\begin{equation*}\label{eq:6}
\mathcal{L} = \mathcal{L}_{\text{CE}} + \alpha \cdot (\mathcal{L}_{\text{MSE}} + \mathcal{L}_{\text{COV}}) +  \beta \cdot \mathcal{L}_{\text{ADV}}
\end{equation*}
where \( \mathcal{L}_{\text{CE}} \) is the cross-entropy loss for classification. We empirically set \( \alpha = 0.01, \beta=0.015 \) and use a 2-layer MLP consistently for fair comparison.


\section{Experiments}

\subsection{Training Data}
In our experiments, we explore using a well-known radiology dataset, \textsc{ROCOv2}~\citep{rückert2024rocov2} to pre-train CLIP models. To introduce adversarial noise in the dataset, we select radiology images from \textsc{MediCaT}~\citep{subramanian-2020-medicat} as target to conduct adversarial image attacking, perturbing clean images from \textsc{ROCOv2}. 

For downstream tasks, we fine-tune pre-trained noisy models on \textsc{ChestXray14} \citep{Wang_2017} for classification, and \textsc{SLAKE}~\citep{liu2021slake} for VQA, respectively. Detailed dataset resources, statistics and examples are provided in Appendix \ref{app:statistics}.

    
    



\subsection{Noisy Model Pre-training}
As shown in \autoref{fig:architecture} (a), we first pre-trian CLIP model (ViT-L/14) on adversarial noisy dataset. The noise ratio $\gamma$ is set to $\{0\%, 5\%, 10\%, 20\%, 30\%\}$, where $0\%$ representing the clean dataset. We randomly select $\gamma$ percentage of image-caption pairs from \textsc{ROCOv2} to attack. To generate image-noisy datasets, we apply adversarial image attack to the selected images. To generate caption-noisy datasets, we perform adversarial caption attack using \textsc{Llama3-8b}, as outlined in \S\ref{sec:noisy-data}.
These noisy models are designed to align radiology images with their corresponding captions, allowing us to analyze the impacts of noise on pre-trained feature extractors, and assess performance differences in downstream tasks. Implementation details on training can be found in Appendix \ref{app:implement}

\subsection{Fine-tuning}
We conduct fine-tuning under three settings: i). \emph{linear probing}~\citep{radford2021learning}, wherein only training a simple linear classifier on top of the frozen features extracted from the noisy models to analyze how upstream noise affects downstream tasks; ii). \emph{MLP-tuning}, where training an MLP classifier without loss regularization; iii). \emph{RAN-finetuning}, using MLP with proposed loss functions.

\begin{figure*}[t]
\centering
\includegraphics[width=0.95\textwidth]{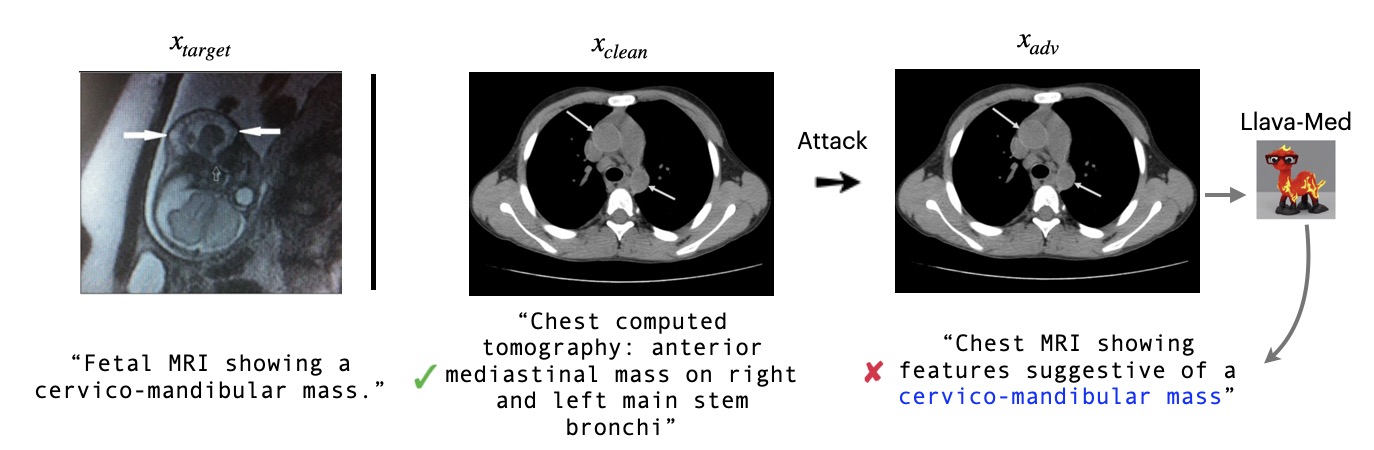}
\caption{An example of generating caption $c_\text{adv}$ of crafted adversarial image $x_\text{adv}$ by black-box VLM, Llava-Med. The default prompt is \emph{“what is the content of this radiology image?”}. \rX{} denotes the generated caption doesn't accurately describe the content of the clean image. \gC{} means otherwise.}
\label{fig:caption-llavamed}
\vspace{-2mm}
\end{figure*}

\paragraph{Chest X-ray Classification}
Given a chest x-ray scan, our fine-tuning objective is to predict possible disease labels from 14 categories\footnote{Details about classification labels are in Appendix \ref{app:statistics} }.

\paragraph{Medical VQA}

To thoroughly evaluate the effectiveness of our method, we then formulate MedVQA as a classification task, where the possible label set consists of all possible answers. Motivated by \citet{dou2022empirical}, we adopt a transformer-based \textit{co-attention} multi-modal fusion module\footnote{Detailed descriptions of medical VQA setting is in Appendix \ref{app:medicalvqa}} that produces cross-modal representations over the image and text encodings, which are then fed to a classifier for predicting the final answer as described in \autoref{fig:architecture}.

\subsection{Evaluation}

\paragraph{Domain Differences} 
To investigate the generalizability of adversarial noisy model comprehensively, we then conduct fine-tuning under two experimental settings: i). a
standard \textbf{in-domain} (ID) setup, in which both the
training and testing data are sourced from the same
dataset; ii). a more challenging \textbf{out-of-domain} (OOD) setup, wherein test data originated from a different dataset.

For ID evaluation, we evaluate the pre-trained noisy models on \textsc{ChestXray14} for chest-xray classification and \textsc{SLAKE} for medical VQA. Under the OOD setting, we use \textsc{CheXpert}~\citep{irvin2019chexpert} and \textsc{VQA-RAD}~\citep{lau2018dataset} respectively. We report performance on both ID and OOD with $\{0\%, 5\%, 10\%, 20\%, 30\%\}$ percentage of downstream datasets. 

\paragraph{Metrics}  All models are evaluated using the macro average of the area under the receiver-operator curve (AUC)~\citep{BRADLEY19971145} and accuracy (ACC) averaged over all labels. 


\section{Results and Analysis}
\label{sec:results}

\subsection{Effectiveness of Adversarial Attack}

In  \autoref{tab:noise}, we evaluate the efficacy of our adversarial attack strategy against white-box models including pre-trained general CLIP~\citep{radford2021learning} and medical CLIP models. We use 5K clean images \( x_{\text{clean}} \) from the \textsc{ROCOv2} validation set and randomly select a targeted images-caption pairs \( (x_{\text{target}}, c_{\text{target}}) \) from \textsc{MediCaT} for each clean image to craft adversarial images \( x_{\text{adv}} \) following the method described in \autoref{fig:adv_img}. We discover that the similarity between \( x_{\text{adv}} \) and \(c_{\text{target}}\), measured by the CLIP score, increases compared to \( x_{\text{clean}} \), which validates the effectiveness of our image attack. In addition, Medical CLIP models are more adept at accurately identifying the content of radiology images (as evidenced by a lower score between \( x_{\text{clean}} \) and \( c_{\text{target}} \)). However, they remain susceptible to our adversarial attack method, which lays the foundation for black-box transferability (See Appendix \ref{app:attack} for details). \autoref{fig:caption-llavamed} shows \textsc{Llava-Med} can be misled to generate inaccurate captions for our adversarially crafted images. 
\begin{table}[t]
\centering
\resizebox{1.0\linewidth}{!}
{%
\begin{tabular}{ccc}
\toprule
Model & Clean Image & Adv. Image  \\
\midrule
CLIP - ViT-L/14 & 0.253 & 0.384 \\
CLIP - Resnet50 & 0.211 & 0.329 \\
PubMedCLIP~\citep{eslami-etal-2023-pubmedclip} & 0.182 & 0.347\\
BioMedCLIP~\citep{zhang2024biomedclip} & 0.174 & 0.312
\\
\bottomrule
\end{tabular}
}
\caption{\textbf{White-box image attacks}. We report the CLIP similarity score between the clean $x_\text{clean}$  or crafted adversarial images $x_\text{adv}$ and the corresponding targeted captions $c_\text{target}$ from \textsc{MediCaT}.}
\label{tab:noise}
\end{table}
\begin{table}[t]
\centering
\vspace{-1mm}
\resizebox{1.0\linewidth}{!}
{%
\begin{tabular}{ccc}
\toprule
Model & Clean Caption & Adv. Caption  \\
\midrule
UniDiffuser~\citep{bao2022one} & 0.431 & 0.274 \\
LLaVA-Med~\citep{li2023llavamed} & 0.565 & 0.392 \\
Mini-GPT4~\citep{zhu2023minigpt4} & 0.493 & 0.287 \\
\bottomrule
\end{tabular}
}
\caption{\textbf{Black-box caption attacks}. We use VLMs to generate a radiology image based on either a clean caption $c_\text{clean}$ or $c_\text{adv}$, and report CLIP score between the generated image (i.e., $\hat{x}_\text{clean}$ and $\hat{x}_\text{adv}$) and $x_\text{clean}$.}
\vspace{-3mm}
\label{tab:black-box-adv-caption}
\end{table}

In \autoref{tab:black-box-adv-caption}, we transfer the crafted adversarial captions to image through advanced VLMs. The similarity between the generated image  $\hat{x}_\text{adv}$ are less similar to the clean image $x_\text{clean}$ than  generated $\hat{x}_\text{clean}$, indicating the effectiveness of caption attack. 


\begin{figure}[t]
\centering
\includegraphics[width=\linewidth]{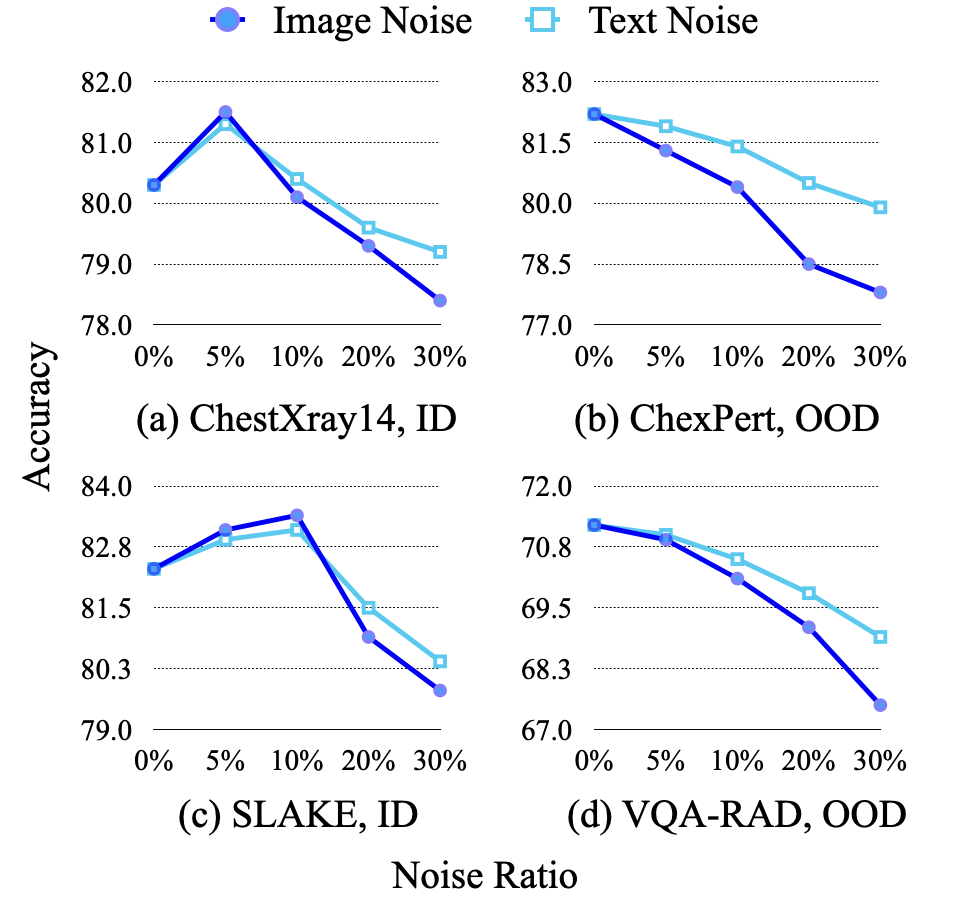}
\caption{\textbf{Linear Probing ID and OOD evaluation} results of CLIP model pre-trained on multi-modal adversarial noise on downstream tasks including chest x-ray classification ((a) and (b)) and medical VQA ((c) and (d)) with various percentages of noise. }
\vspace{-4mm}
\label{fig:result_noise}
\end{figure}
\subsection{Adversarial Noise Evaluation}
\label{sec:noise_evaluation}
\begin{figure*}[t]
    \centering
    \includegraphics[width=1\textwidth]{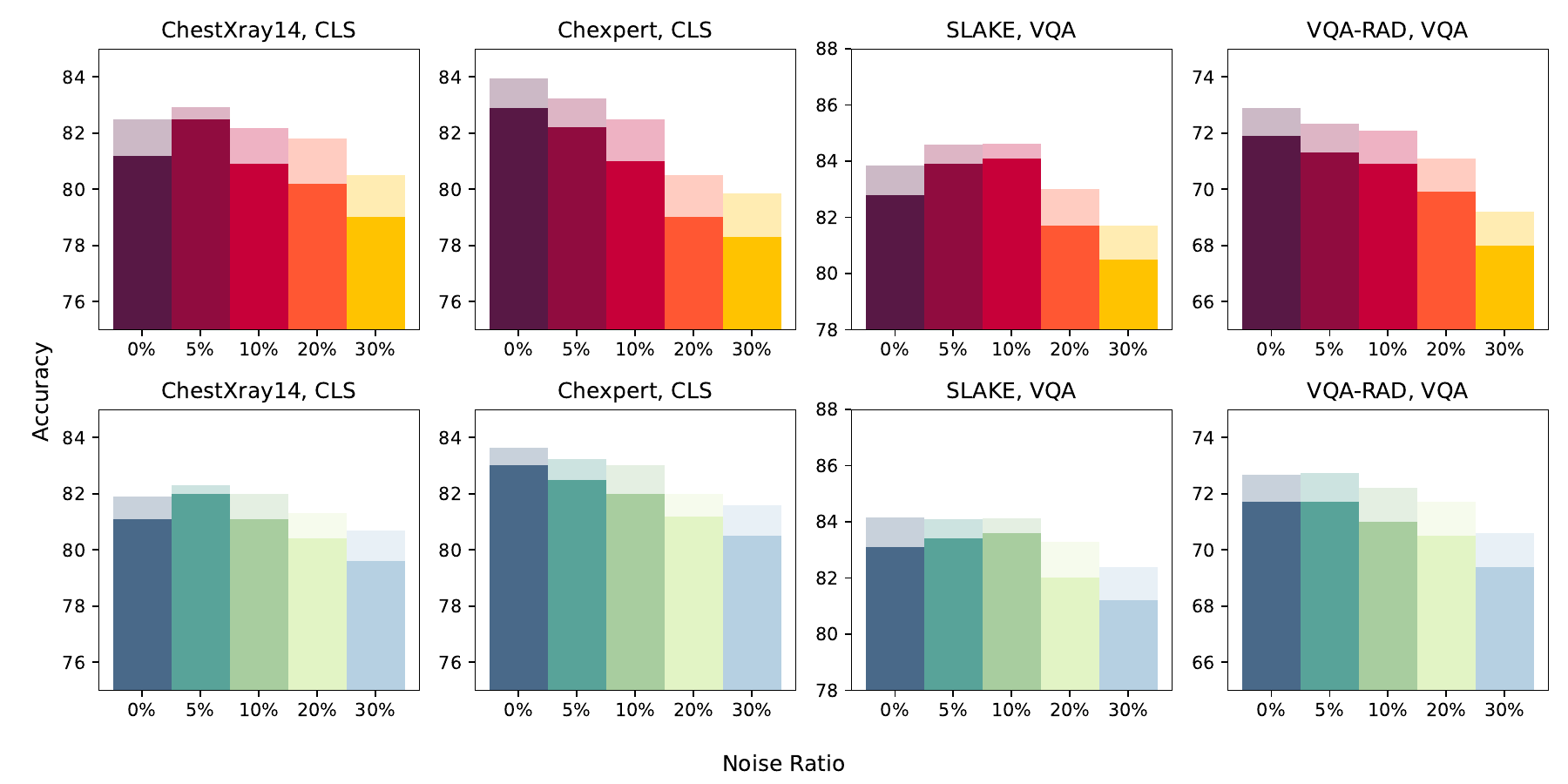}
    \caption{\textbf{Evaluation of \textsc{RAN} fine-tuning} on ID and OOD downstream tasks, compared to MLP tuning. We use CLIP models pre-trained on noisy \textsc{ROCOv2} dataset with, [\textbf{first row}]: adversarial images; and [\textbf{second row}]: adversarial captions.  The improvements of RAN are presented by stacked bars with light colors.}
    \vspace{-3mm}
    \label{fig:result-mitigate}
\end{figure*}

To explore the effects of adversarial multi-modal noise from upstream training on downstream tasks, we show the accuracy of evaluating pre-trained noisy models on both ID and OOD tasks across all noise ratios, under linear probing setting, in \autoref{fig:result_noise}. We empirically reveal the following insights: 

\textbf{Introducing a moderate level of noise, such as 5\% or 10\%, during pre-training can actually improve a model's robustness and performance on ID downstream tasks.} We hypothesize this slight noise acts as a form of regularization, helping the model generalize better to similar data seen during fine-tuning. However, increasing the noise beyond this threshold starts to degrade the model's performance, leading to poorer results. This finding aligns with \citet{song2022learning}, suggesting a balance in noise levels is critical for optimal model training, as excessive noise can introduce too much variability.

{\textbf{The performance on OOD downstream tasks consistently diminishes as the noise level in pre-training increases.} High levels of noise make it harder for the model to adapt to new and unseen data, reducing its ability to generalize effectively beyond the training domain.

\textbf{Image attacks tend to be more potent than caption attacks in affecting model performance.} Specifically, across all four tasks, models subjected to image noise exhibit more significant changes than those exposed to text noise. This suggests that image perturbations can disrupt the model’s internal representations more effectively, leading to greater performance improvement or degradation. 
\begin{table}[]
\resizebox{1.0\linewidth}{!}
{%
\begin{tabular}{ccc}
\toprule
 $\gamma$  & Setting   &  \textsc{Chest-Xray14}    \\
\midrule
0  & Base         & 80.3 \\
\multirow{3}{*}{5}  & Ours (LP)    & 81.3 \\
   & Random (LP)  & 80.6 \\
   & Random + RAN & 81.5 \\
\multirow{3}{*}{10} & Ours (LP)    & 80.4 \\
   & Random (LP)  & 79.8 \\
   & Random + RAN & 80.8 \\
\multirow{3}{*}{20} & Ours (LP)    & 79.6 \\
   & Random (LP)  & 79.1 \\
   & Random + RAN & 80.2 \\
\multirow{3}{*}{30} & Ours (LP)    & 78.9 \\
   & Random (LP)  & 78.2 \\
   & Random + RAN & 79.4 \\
   \bottomrule
\end{tabular}
}
\caption{Performance Comparison with Random Noise}
\label{tab:random}
\end{table}
\begin{table}[t]
\centering
\resizebox{1.0\linewidth}{!}
{%
\begin{tabular}{llllll}
\toprule 
Noise Type & $\gamma$ & \multicolumn{2}{c} {\textsc{ChestXray14}} & \multicolumn{2}{c} {\textsc{CheXpert}} \\
 & \multicolumn{1}{l}{} & \multicolumn{1}{c}{AUC} & \multicolumn{1}{c}{ACC} & \multicolumn{1}{c}{AUC} & \multicolumn{1}{c}{ACC} \\
 \midrule
\multirow{5}{*}{Image} 
 & 0 &  65.2 & 72.4 & 68.6 & 76.1 \\
 & 5 & 65.5 $\uparrow$ & 72.9 $\uparrow$ & 69.5 $\uparrow$ & 76.8 $\uparrow$ \\
 & 10 & 64.7 & 71.2 & 69.3 $\uparrow$ & 76.3 $\uparrow$ \\
 & 20 & 64.1 & 70.6 & 67.8  & 75.4 \\
 & 30 & 63.5 & 69.9 & 67.1 & 74.9 \\ 
\hdashline
\multirow{5}{*}{Caption} 
 & 5 & 65.8 $\uparrow$ & 72.8 $\uparrow$ & 69.0 $\uparrow$  & 76.4 $\uparrow$ \\
 & 10 & 65.4 $\uparrow$ & 72.2 & 68.4  & 76.0 \\
 & 20 & 64.7 & 71.9 & 68.3 & 75.8  \\
 & 30 & 64.1 & 71.2 & 67.6 & 75.2 \\
 \bottomrule
\end{tabular}
}
\caption{\textbf{Zero-shot Evaluations} on Chest X-ray classification tasks across different pre-trained noise ratios ($\gamma$). ($\uparrow$) indicates improvements from the clean baseline.}
\label{tab:zero-shot}
\end{table}

\textbf{Comparing with Random Noise}

To disentangle the two factors (how downstream performance differs between our crafted adversarial caption noise, and just random noise?), we also test additional baselines that introduce small amount of random gaussian noise to caption input (\autoref{tab:random}). As expected, when noise is 5\%, random noise has a slightly lesser improvement on downstream performance compared to our crafted noise; as noise increases, random noise deteriorates performance more rapidly. Nevertheless, our study confirms that our noise mitigation strategy, RAN, is effective not only against our crafted adversarial noise but also against random noise.


\subsection{Zero-shot Evaluation}
To comprehensively study the robustness of pre-trained noisy models without fine-tuning, we perform a zero-shot chest x-ray classification on two datasets: \textsc{ChestXray14} and \textsc{CheXpert}. To match text with the encoded image embeddings, we use prompts of \emph{\{label\}} and \emph{No \{label\}} (e.g., "Atelectasis" vs. "No atelectasis") following \citet{You_2023}. The results are illustrated in \autoref{tab:zero-shot}.

Following the findings from \S\ref{sec:noise_evaluation}, introducing slight noise enhances pre-trained model robustness and performs better on zero-shot classification tasks, while excessive noise in turn hurts the performance.

\subsection{Effectiveness of RAN Fine-tuning}
\autoref{fig:result-mitigate} shows the effect of our proposed RAN fine-tuning on mitigating upstream adversarial noise. To disentangle two possible reasons--\emph{RAN regularization} or \emph{extra parameters from MLP}--improve models' robustness, we compare it against baselines using MLP-tuning.  

From experimental results, incorporating RAN enhances overall performance across all ID and OOD tasks against both image and caption attacks. The performance improvement observed with a 5\% noise ratio in the ID \textsc{ChestXray} task and a 10\% noise ratio in the \textsc{SLAKE} task under linear probing setting is less pronounced with RAN fine-tuning, indicating its effectiveness in rectifying the influences of noise. Particularly for OOD tasks, the improvement is slightly more significant. Moreover, we notice that the improvement on caption-noisy models is less significant than image-noisy models when noise starting to degrade model performance, potentially because the impact of image noise on feature extractors is greater, and RAN rectify such effects during fine-tuning (See Appendix \ref{tab:full_results} for full results).

\begin{table}[t]
\centering
\resizebox{1.0\linewidth}{!}
{%
\begin{tabular}{cccccc}
\toprule
Model &  Setting & SLAKE & VQA-Rad \\
\midrule
\multirow{2}{*}{\textsc{BioMedCLIP}} & Baseline & 88.9  & 79.8    \\ & with RAN  & 90.2  & 80.9    \\
\multirow{2}{*}{\textsc{PubMedCLIP}} & Baseline & 82.5  & 80.0    \\ & with RAN  & 83.1  & 80.5    \\
\multirow{2}{*}{\textsc{LLaVa-Med}}  & Baseline & 84.2  & 85.3    \\ & with RAN  & 85.1  & 85.2  \\
\multirow{2}{*}{\textsc{PMC-CLIP}}   & Baseline & 88.0  & 84.0    \\  & with RAN  & 89.7  & 84.8   \\
\bottomrule
\end{tabular}
}
\caption{Comparison with other baseline VLMs.}
\vspace{-2mm}
\label{tab:baseline}
\end{table}
\begin{table}[]
\vspace{-2mm}
\centering
\resizebox{0.5\textwidth}{!}{%
\begin{tabular}{cccc}
\toprule
        $\gamma$            &  Setting       & \textsc{Chest-Xray} & \textsc{Chexpert} \\
        \midrule
\multirow{3}{*}{5}  & LP     & 81.5       & 81.3     \\
                    & RAN    & 82.9       & 83.2     \\
                    & NMTune & 82.6       & 82.7     \\
\multirow{3}{*}{10} & LP     & 80.1       & 80.4     \\
                    & RAN    & 82.2       & 82.5     \\
                    & NMTune & 81.5       & 81.6     \\
\multirow{3}{*}{20} & LP     & 79.3       & 78.5     \\
                    & RAN    & 81.8       & 80.5     \\
                    & NMTune & 80.7       & 79.6     \\
\multirow{3}{*}{30} & LP     & 78.2       & 77.8     \\
                    & RAN    & 80.5       & 79.8     \\
                    & NMTune & 79.6       & 78.7    \\
                    \bottomrule
\end{tabular}
}
\caption{Comprison with NMTune.}
\label{tab:nmtune}
\end{table}
Given that SOTA medical VLMs are generally pre-trained on large amounts of data (e.g., BioMed-CLIP pretrained on 15M private medical image-text pairs), we do not believe it's fair to directly compare the finetuned results. However, how our noise mitigation strategy RAN would fare on other SOTA VLMs is indeed an interesting question. Fine-tuning such VLMs with RAN can test if RAN is resilient against various types of noise, as such VLMs might pre-trained with unknown noise, we present the following results with finetuning SOTA VLMS on medical VQA tasks, with our proposed RAN as noise mitigation:
As shown in \autoref{tab:baseline}, we can see that with RAN, almost all SOTA VLMs exhibit better performance across the board, which validates the effectiveness of our proposed noise mitigation fine-tuning strategy. With only PubMedCLIP has less improvement. We hypothesize this is because PubMedCLIP was pre-trained on a small-scale, high-quality dataset, whereas the others were pre-trained on larger-scale datasets, which may contain more noise.

In \autoref{tab:nmtune}, we present the experimental results on chest classification task between applying RAN and another noise model mitigation approach NMTune~\citep{chen2024understanding}. Our method performs better on rectifying adversarial noise than NMTune in the above case. We hypothesize it’s because NMTune tries to mitigate label noise, and focuses on rectifying the features shaped by such noise, whereas our adversarial loss regularization mainly focuses on against adversarial noises.

\paragraph{Ablations}

We perform extensive ablations to show that every component of RAN benefits the overall system (\autoref{tab:ablation}). 
The results indicate that our proposed \(\mathcal{L}_\text{COV}\) and \(\mathcal{L}_\text{ADV}\) effectively mitigate the impact of adversarial image noise. While the improvement from using only \(\mathcal{L}_\text{MSE}\) is relatively modest, \(\mathcal{L}_\text{ADV}\) and \(\mathcal{L}_\text{COV}\) shows limited enhancement in performance for models with clean upstream data ($\gamma =0$) compared to noisy models ($\gamma = 20$). We hypothesize that this is because \(\mathcal{L}_\text{ADV}\) and \(\mathcal{L}_\text{COV}\) are primarily designed to address feature changes induced by upstream noise, which may not provide significant benefits in clean datasets. Combining all loss terms proposed in RAN effectively improves performance against MLP-tuning.
\begin{table}[t]
\centering
\resizebox{1.0\linewidth}{!}
{%
\begin{tabular}{cccccc}
\toprule
$\mathcal{L}_\text{MSE}$ & $\mathcal{L}_\text{COV}$ & $\mathcal{L}_\text{ADV}$ & $\gamma$ & \textsc{ChestXray14} & \textsc{SLAKE} \\
\midrule
\rX{} & \rX{} & \rX{} & 0 &  81.2 & 82.8  \\
\gC{} & \rX{} & \rX{} & 0 & 81.2 & 83.0  \\
\rX{} & \gC{} & \rX{} & 0 &  81.7 & \hlc[violet!10]{83.3} \\
\rX{} & \rX{} & \gC{} & 0 & 81.4 & {83.2} \\
\gC{} & \gC{} & \rX{} & 0 &  \hlc[violet!10]{81.9} & \hlc[violet!10]{83.4}  \\
\gC{} & \gC{} & \gC{} & 0 &  \hlc[violet!10]{82.5} & \hlc[violet!10]{83.8} \\
\hdashline
\rX{} & \rX{} & \rX{} & 20 & 80.2  & 81.8  \\
\gC{} & \rX{} & \rX{} & 20 & 80.5  &  82.1 \\
\rX{} & \gC{} & \rX{} & 20 & \hlc[violet!10]{80.8} & \hlc[violet!10]{82.3}  \\
\rX{} & \rX{} & \gC{}  & 20 & \hlc[violet!10]{80.9} & 82.1  \\
\gC{} & \gC{} & \rX{} & 20 & \hlc[violet!10]{81.3} & \hlc[violet!10]{82.7} \\
\gC{} & \gC{} & \gC{} & 20 & \hlc[violet!10]{81.8}  & \hlc[violet!10]{83.0} \\
\bottomrule
\end{tabular}
}
\caption{\textbf{Ablation Study} on loss terms of ID tasks with image attack. $\gamma$ denotes noise ratio in pre-trained dataset. \hlc[violet!10]{Highlighted} denotes improvement $\geq 0.5$.}
\vspace{-4mm}
\label{tab:ablation}
\end{table}




\section{Conclusion}

Despite the success development of medical VLMs, most such models are vulnerable to adversarial attack and still lag behind in transferring to downstream tasks robustly. In this work, we discuss how upstream adversarial noise affects various medical downstream tasks by crafting adversarial multi-modal medical samples. Through extensive experiments, we found that even minor adversarial noise in pre-training datasets can enhance ID performance while degrading OOD generalization. To this end, we introduce a light-weight fine-tuning recipe, RAN, effectively mitigating noise effects by refining the feature space and enforcing robust margins to defend adversarial noise. 

\section{Limitations}
Several limitations restrict the scope of our work. To begin, our choice of downstream tasks--chest X-ray classification and medical VQA tasks--is nonexhaustive, and it is possible that our findings would not generalize well for the broad spectrum of medical applications. Given that medical datasets can be quite limited compared to other general datasets due to its private nature, pretrain a VLMs are also limited. Another restriction is that we only attacked radiology images and their captions, offering a glimpse into possible vulnerabilities but not a complete picture. This means our findings may not apply to other kinds of medical images or related text data. Expanding to various medical imaging and datasets in future work will be crucial for more comprehensive insights and real-world applicability. 

Other potential avenues for exploration entail different noise type and evaluate the nature of how noise shapes pre-trained features can be useful. Future work should explore optimizing noise levels and further enhancing the robustness of VLMs to various adversarial scenarios to maintain high performance across diverse medical domains 

\section*{Acknowledgment}
This work is partially supported by NSF NAIRR240016, NIH R21EB034911, and Google Cloud research credits. 


\bibliography{custom}
\appendix

\section*{Appendix }
\label{sec:appendix}

\section{Multi-modal Adversarial Attack}
\label{app:attack}
\autoref{tab:noise} validates the effectiveness of white-box attack against CLIP models. In
\autoref{tab:black-box-adv}, we transfer the crafted adversarial examples in order to evade large VLMs and mislead them into generating targeted responses. The similarity between the generated response  $c_\text{adv}$ are more similar to the targeted text $c_\text{target}$ than  $c_\text{clean}$, indicating the effectiveness of our method towards advanced large VLMs.

\begin{table}[h]
\centering
\vspace{-1mm}
\resizebox{1.0\linewidth}{!}
{%
\begin{tabular}{ccc}
\toprule
Model & Clean image & Adv. image  \\
\midrule
UniDiffuser~\citep{bao2022one} & 0.287 & 0.594 \\
LLaVA-Med~\citep{li2023llavamed} & 0.246 & 0.483\\
\bottomrule
\end{tabular}
}
\caption{\textbf{Black-box image attacks}. We report CLIP score between the generated caption of input images (i.e., $x_\text{clean}$ or crafted $x_\text{adv}$) and targeted caption $c_\text{target}$,}
\vspace{-3mm}
\label{tab:black-box-adv}
\end{table}



\section{Training}
\subsection{Datasets}
\label{app:statistics}

\paragraph{ROCOv2~\citep{rückert2024rocov2}} provides 79,789 radiological images with associated captions and medical concepts. The image–text pairs are captured from PubMed~\citep{} articles.  It is an updated version of the ROCO~\citep{roco2018pelka} dataset published in 2018, and adds 35,705 new images added to PMC since 2018. 

\paragraph{MediCaT~\citep{subramanian-2020-medicat}} includes medical images, captions, subfigure-subcaption annotations, and inline textual references from. It consists of 217,060 figures from 131,410 open access papers, 7,507 subcaption and subfigure annotations for 2,069 compound figures.

\paragraph{ChestXray14~\citep{Wang_2017}} is a medical imaging dataset that includes 112,120 frontal-view X-ray images from 30,805 unique patients, collected between 1992 and 2015. It features fourteen common disease labels extracted from radiological reports. The disease categories are: \emph{Atelectasis, Cardiomegaly, Effusion, Infiltrate, Mass, Nodule, Pneumonia, Pneumothorax, Consolidation, Edema, Emphysema, Fibrosis, Pleural Thickening, Hernia, No finding}.

\paragraph{CheXpert~\citep{irvin2019chexpert}} is a large dataset of chest X-rays of 65,240 patients, with 14 observation labels collected from Stanford Hospital. The included 14 labels are: \emph{Enlarged Cardiom, Cardiomegaly, Lung Lesion, Lung Opacity, Edema, Consolidation, Pneumonia, Atelectasis, Pneumothorax, Pleural Effusion, Pleural Other, Fracture, Support Devices and No Finding}.
\begin{figure}[h]
\centering
\includegraphics[width=\linewidth]{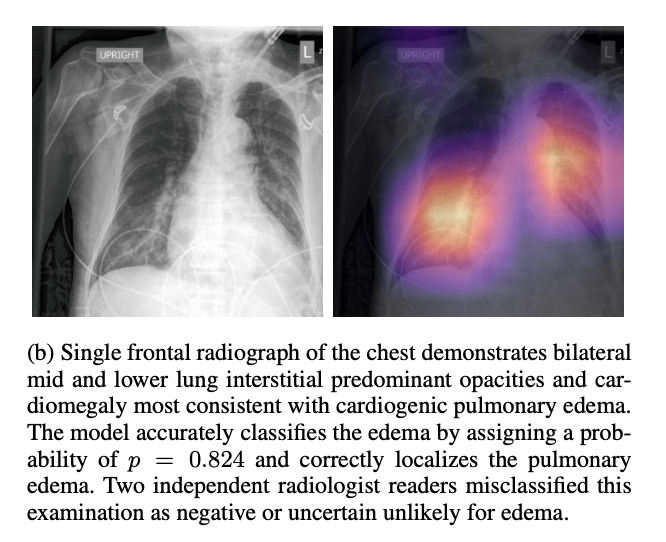}
\caption{An example of image-caption in CheXpert dataset.}
\label{fig:chest}
\end{figure}

\paragraph{SLAKE~\citep{liu2021slake}}~ a
bilingual Med-VQA benchmark containing 6428 radiology images (X-rays, CT scans, MRIs) and 14,028 question-answer pairs. It includes both "closed-ended" questions, and more challenging "open-ended" questions. Fo simplicity, we only report performance evaluated on "closed-ended" questions. An example image-question pair is shown in \autoref{fig:vqa-slake}.

\begin{figure}[h]
\centering
\includegraphics[width=\linewidth]{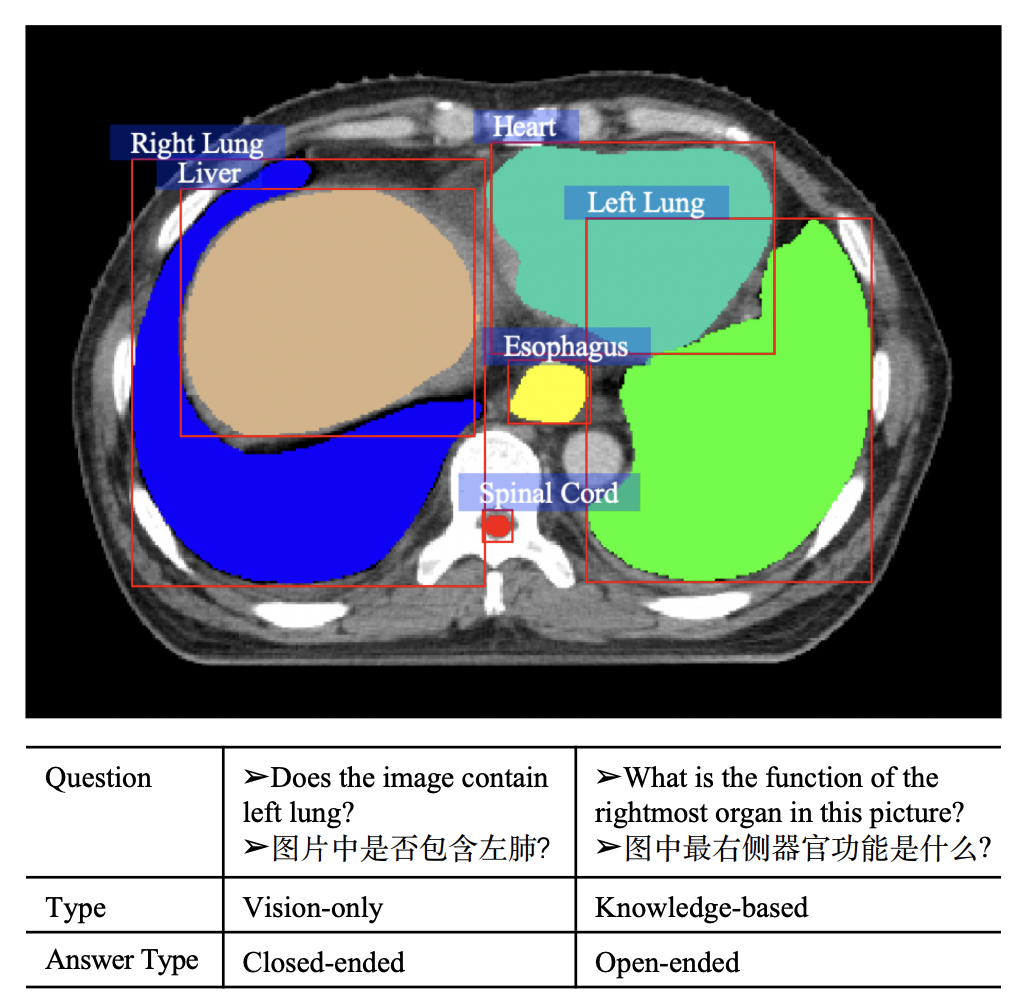}
\caption{An example of image-question in SLAKE dataset.}
\label{fig:vqa-slake}
\end{figure}

\paragraph{VQA-RAD~\citep{lau2018dataset}} contains 3515 question–answer pairs generated by clinicians and 315 radiology images that are evenly distributed over the head, chest, and abdomen. Each image is associated with multiple questions. Half of the answers are closed-ended (i.e., yes/no type), while the rest are open-ended with either one-word or short phrase answers.

\begin{table}[h]
\vspace{-2mm}
\centering
\resizebox{0.5\textwidth}{!}{%
\begin{tabular}{llccc}
\toprule
\textsc{Dataset} & \textsc{Split} & \multicolumn{1}{l}{\textsc{Image \#}} & \textsc{\begin{tabular}[c]{@{}c@{}}Caption \#\\ / Question \# \end{tabular}} & \textsc{Answer \#}\\
\midrule

\multirow{3}{*}{\textsc{ROCOv2}} 
& \textsc{Train} & 59,958 & 59,958 & / \\
& \textsc{Valid} & 9904 & 9904 & / \\
& \textsc{Test} & 9927 & 9927 & / \\
\midrule

\multirow{3}{*}{\textsc{MediCaT}} & \textsc{Train} & 141,089 & 141,089 & / \\
& \textsc{Valid} & 32,559 & 32,559 & / \\
& \textsc{Test} & 43,412 & 43,412 & / \\
\midrule

\multirow{3}{*}{\textsc{ChestXray14}} & \textsc{Train} & 78,484  & / & / \\
& \textsc{Valid} & 11,212 & / & / \\
& \textsc{Test} & 22,424 & / & / \\
\midrule
\multirow{3}{*}{\textsc{CheXpert}} & \textsc{Train} & 224,316 & / & / \\
& \textsc{Valid} & 235 & / & / \\
& \textsc{Test} & 669 & /  & / \\
\midrule
\multirow{3}{*}{\textsc{SLAKE}} 
& \textsc{Train} & 450 & 9,849 & 9,849 \\
& \textsc{Valid} & 96 & 2,109 & 2,109 \\
& \textsc{Test} & 96 & 2,070 & 2,070 \\
\midrule
\multirow{3}{*}{\textsc{VQA-RAD}} & \textsc{Train} & \multirow{2}{*}{315} & 3,064 & 3,064  \\
& \textsc{Test} &  & 451 & 451 \\
\bottomrule

\end{tabular}
}
\caption{ Medical Dataset Statistics.}
\label{tab:dataset}
\vspace{-3mm}
\end{table}

\begin{table*}[ht]
\centering
\resizebox{0.8\textwidth}{!}
{%
\begin{tabular}{ccccccc}
\toprule
\textsc{Noise} Type & $\gamma$ & \textsc{Classifier} & \textsc{ChestXray14} & \textsc{CheXpert} & \textsc{SLAKE} & \textsc{VQA-RAD} \\
\midrule
\multirow{15}{*}{Image} & \multirow{3}{*}{0} 
 & LP & 80.3 & 82.2 & 82.3 & 71.2 \\
 &  & MLP & 81.2 & 82.9 & 82.8 & 71.9 \\
 & \multirow{3}{*}{5}
 & LP & 81.5 & 81.3 & 83.1 & 70.9 \\
 &  & MLP & 82.5 & 82.2 & 83.9 & 71.3 \\
 &  & RAN & 82.9 & 83.2 & 84.6 & 72.3 \\
 & \multirow{3}{*}{10} 
 & LP & 80.1 & 80.4 & 83.4 & 70.1 \\
 &  & MLP & 80.9 & 81.0 & 84.1 & 70.9 \\
 &  & RAN & 82.2 & 82.5 & 84.6 & 72.1 \\
 & \multirow{3}{*}{20} 
 & LP & 79.3 & 78.5 & 80.9 & 69.1 \\
 &  & MLP & 80.2 & 79.0 & 81.8 & 69.9 \\
 &  & RAN & 81.8 & 80.5 & 83.0 & 71.1 \\
 & \multirow{3}{*}{30} 
 & LP & 78.2 & 77.8 & 79.8 & 67.5 \\
 &  & MLP & 79.0 & 78.3 & 80.5 & 68.0 \\
 &  & RAN & 80.5 & 79.8 & 81.7 & 69.2 \\
 \midrule
\multirow{15}{*}{Caption} & \multirow{3}{*}{0} 
 & LP & 80.3 & 82.2 & 82.3 & 71.2 \\
 &  & MLP & 81.1 & 83.0 & 83.1 & 71.7 \\
 &  & RAN & 81.9 & 83.6 & 84.2 & 72.7 \\
 & \multirow{3}{*}{5} 
 & LP & 81.3 & 81.9 & 82.9 & 71.0 \\
 &  & MLP & 82.0 & 82.5 & 83.4 & 71.7 \\
 &  & RAN & 82.3 & 83.2 & 84.1 & 72.7 \\
 & \multirow{3}{*}{10} 
 & LP & 80.4 & 81.4 & 83.1 & 70.5 \\
 &  & MLP & 81.1 & 82.0 & 83.6 & 71.0 \\
 &  & RAN & 82.0 & 83.0 & 84.1 & 72.2 \\
 & \multirow{3}{*}{20} 
 & LP & 79.6 & 80.5 & 81.5 & 69.8 \\
 &  & MLP & 80.4 & 81.2 & 82.0 & 70.5 \\
 &  & RAN & 81.3 & 82.0 & 83.3 & 71.7 \\
 & \multirow{3}{*}{30} 
 & LP & 78.9 & 79.8 & 80.4 & 68.9 \\
 &  & MLP & 79.6 & 80.5 & 81.2 & 69.4 \\
 &  & RAN & 80.7 & 81.6 & 82.4 & 70.6\\
 \bottomrule
\end{tabular}
}
\caption{The full results of fine-tuning with  linear probing , MLP and RAN across all noise ratios. }
\label{tab:full_results}
\end{table*}

\subsection{Pre-training with CLIP models}
\paragraph{CLIP Contrastive Loss}
\begin{equation}
\ell_{u \rightarrow v}^{i} = -\log \frac{\exp(\text{sim}(u_i, v_i)/\tau)}{\sum_{j=1}^N \exp(\text{sim}(u_i, v_j)/\tau)}
\end{equation}
where \( u \) and \( v \) are the normalized vectors from the image and text encoders, respectively. \( (u_i, v_i) \) is a positive pair, \(\text{sim}\) is a function that calculates the similarity between the vectors, and \( \tau \) is the learnable temperature parameter. \( N \) represents the mini-batch size of image-text pairs. \( \ell_{u \rightarrow v}^{i} \) denotes the InfoNCE loss from image \( i \) to the texts, while \( \ell_{v \rightarrow u}^{i} \) represents the loss in the opposite direction. The final loss in CLIP is defined as:

\begin{equation}
L_{\text{CLIP}} = \frac{1}{2N} \sum_{i=1}^N (\ell_{u \rightarrow v}^{i} + \ell_{v \rightarrow u}^{i})
\end{equation}

\subsection{Medical VQA}
\label{app:medicalvqa}
Given a MedVQA training dataset denoted as \( T = \{(v_i, q_i, a_i)\}_{i=1}^V \) of size \( V \), where \( v_i \) is a medical image, \( q_i \) is the corresponding natural language question, and \( a_i \) is the natural language answer, our objective is to learn to generate the correct answer \( a_i \) for a given image-question pair \((v_i, q_i)\). The features obtained from the image and question encoder are concatenated as \( f_v(v_i) \oplus f_t(q_i) \)
We then formulate MedVQA as a multi-label classification function \( F : \mathbb{R}^n \times \mathbb{R}^{m \times l} \rightarrow \{0, 1\}^{|A|} \), where \( A \) is the overall set of possible answers and \( F(f_v, f_q) = a_i \) for the one-hot encoded answer \( a_i \).

\paragraph{Multi-modal Fusion Module}
\autoref{fig:co-attention} presents the co-attention architecture we use to fuse visual and text features.

\begin{figure}[ht]
\centering
\includegraphics[width=0.8\linewidth]{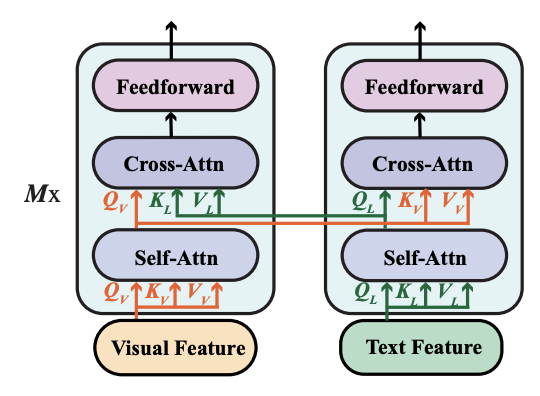}
\caption{Illustration of multi-modal fusion modules through co-attention architecture.}
\label{fig:co-attention}
\end{figure}

\subsection{Implementation Details}  
\label{app:implement}

Our implementation is based on OpenCLIP~\citep{cherti2022reproducible}. We utilize the CLIP with ViT-L/14 architecture, with input images at a resolution of 240. The model comprises a total of 24 layers, which are divided into 4 stages, each encompassing 6 layers. The CLIP model has been pre-trained on the \textsc{DataComp-1B} dataset~\citep{gadre2023datacomp} to ensure robust image-text matching in general domains, which facilitates effective fine-tuning on the medical domain where data is more limited. Following \citet{zhang2024biomedclip}, we use the Adam optimizer with \(\beta_1 = 0.9 \text{ and } \beta_2 = 0.98\), a cosine decay learning rate scheduler with an initial value of 5e-4 at batch size of 16, and the warm-up step set to 2000, conducting 30 epochs for training on 4 A40 GPU.

\section{Full Results of Fine-tuning}
Results are presented in \autoref{tab:full_results}.

\end{document}